# Phrase Pair Mappings for Hindi-English Statistical Machine Translation

**Sreelekha S, Pushpak Bhattacharyya**

Indian Institute of Technology (IIT) Bombay, India

{sreelekha, pb}@cse.iitb.ac.in

**Abstract**

In this paper, we present our work on the creation of lexical resources for the Machine Translation between English and Hindi. We describes the development of phrase pair mappings for our experiments and the comparative performance evaluation between different trained models on top of the baseline Statistical Machine Translation system. We focused on augmenting the parallel corpus with more vocabulary as well as with various inflected forms by exploring different ways. We have augmented the training corpus with various lexical resources such as lexical words, synset words, function words and verb phrases. We have described the case studies, automatic and subjective evaluations, detailed error analysis for both the English to Hindi and Hindi to English machine translation systems. We further analyzed that, there is an incremental growth in the quality of machine translation with the usage of various lexical resources. Thus lexical resources do help uplift the translation quality of resource poor languages.

## 1. Introduction

The quality of the Machine Translation (MT) is defined by how well the morphological inflections and the linguistic properties are being transferred (Kunchukuttan et.al., 2012; Ramananthan et. al., 2011; Dorr et. al., 1994; Och et. al., 2001; Ooch et. al., 2003; Knight K. 1999). Linguistic resources can play a major role to cover the various linguistic phenomena during MT. Many ongoing MT system developments are there for Indian languages using rule-based as well as statistical-based approaches (Antony P. J. 2013; Ashan et. al., 2010; Brown et. al., 1993; Nair, et.al., 2012; Sreelekha et. al., 2013; Sreelekha et. al., 2015; Sreelekha et. al., 2015; Sreelekha et. al., 2017; Sreelekha et. al., 2018). In this paper, we discuss various approaches used in English to Hindi Statistical MT system and vice versa to improve the quality of machine translation. Hindi is morphologically very complex compared to English with its linguistic diversity.

Consider the English sentence,
    *English - Then umbrella might also be needed.*
The English-Hindi SMT system translated it as,
    *Hindi-* तब छतरी पड़ सकता। *{tab chathari pad sakta}*
    *{Then umbrella needs}*

Here the system fails to translate the English verb phrase "*might also be needed*" properly and it translated a part wrongly as *"पड़ सकता"{pad sakta}* and *is failed to translate to its correct Hindi translation* "**की आवश्यकता भी पड़ सकती है**" {*ki aavasyakath bhi pad sakti hae*}{*might also be needed*}.

    Lexical resources can play a major role to learn the various inflected forms during this kind of situations. If we are able to train the machine with the verb phrase translation of "*might also be needed*" as "**की आवश्यकता भी पड़ सकती है**" {*ki aavasyakath bhi pad sakti hae*} then it will help the machine to learn the inflections correctly.

## 2. Related Work

Statistical models will get easily affected by the word order, since SMT works based upon the source-target word alignments (Brown et. al., 1993; Kunchukuttan et.al., 2012; Ramananthan et. al., 2011; Dorr et. al., 1994; Och et. al., 2001; Ooch et. al., 2003; Knight K. 1999). English follows SVO order, on the other hand Hindi follows SOV order. In addition, Hindi has post-position suffixes, which are pre-position prepositions in the case of English. Moreover there are challenges of ambiguities such as; Lexical ambiguity, Structural ambiguity and Semantic ambiguity. In this kind of scenario, usage of lexical resources will help the MT system to learn the word order and inflected forms. In addition various categories of word forms such as lexical words, verb phrases, semantic words, morphological forms etc will help to handle the ambiguity to a great extend. Improving th equality of Mt output by adding dictionary words to the corpus was studied by Och and Ney in their paper (Och and Ney, 2003). We have explained the extraction of various lexical resources, it's validation, time it took to create the resource and its augmentation process in machine translation in the experimental Section. The comparative performance analysis with phrase based model with that of augmented lexical resources is described in Section 3 & 4.

## 3. Experimental Discussion

We discuss the various experiments conducted on our English-Hindi and Hindi-English Baseline SMT system by augmenting various lexical resources and the comparisons of results in the form of an error analysis. We have used Moses (Koehn et. al., 2007) and Giza++[1] for modeling the baseline system. Table 1 shows the statistics of corpus and the various lexical resources used for our experiments. The lexical resources include

---
[1] http://www.statmt.org/

programmatically extracted entries as well as manually created entries. The extracted entries have been validated manually. The lexical resources has been created and validated by two English-Hindi bilingual experts over a period of 2 years with qualifications of Master degree in Hindi and English Literature respectively. The experiments conducted are as follows: Baseline SMT system with an uncleaned corpus, Baseline SMT system with a cleaned corpus, Baseline SMT system with IndoWordnet extracted words, Baseline SMT system with Suffix splitted corpus, Baseline SMT system with Function words and Baseline SMT system with verb phrases. The results are shown in Tables 2, 3, 4 and 5. The detailed description of each experiment is explained with an example as listed below:

| Sl. No | Corpus Source | Training Corpus [Manually cleaned and aligned] | Corpus Size [Sentences] |
|---|---|---|---|
| 1 | ILCI | Tourism | 24250 |
| 2 | ILCI | Health | 24250 |
| 3 | DIT | Tourism | 20000 |
| 4 | DIT | Health | 20000 |
| | | Total | 88500 |

| Sl. No | Lexical Resource Source | Lexical Resources in Corpus | Lexical Resource Size [Words] |
|---|---|---|---|
| 1 | CFILT, IIT Bombay | IndoWordnet Synset words | 200000 |
| 2 | CFILT IIT B | Function Words | 15000 |
| 3 | CFILT IIT B | Verb Phrases | 85000 |
| | | Total | 300000 |

| Sl. No | Corpus Source ILCI | Tuning corpus size [Sentences] | Testing Corpus Size [Sentences] |
|---|---|---|---|
| 1 | Touris, Health | 500 | 1000 |

**Table 1: Statistics of Corpus and Lexical Resources Used**

### 3.1 Baseline system with an unclean corpus

The corpus used for our experiments are taken from ILCI corpus in Tourism and Health domain. The corpus was having mis-alignments, wrong and missing translations which affected the quality of translation. Consider a sentence from the uncleaned English-Hindi corpus, where the translation is wrong,

**Hindi** : गर्मी से लू लगने से सिर दर्द तथा भूख न लगना.*{garmi se loo lagne se sir dard tadha bhookh na lagna}{Headache and lack of appetite because of sunstroke in summer.}*
**Wrong English Translation**:
*Summer headache and lack of appetite*
Here, the above English translation is wrong and the trained models generated with this uncleaned corpus also results in poor quality translation. The experiemntal results with uncleaned corpus is shown in the Table 2, 3, 4 and 5. We observed that the quality of the parallel corpus can help in generating the good quality translation models. Hence we focussed on cleaning the parallel corpus before training.

### 3.2 Baseline system with cleaned corpus

The English-Hindi bilingual experts have cleaned the parallel corpus such as, removed the unwanted characters and wrong translations and also corrected the missing translations and phrases. We have manually aligned the source and target sentences in the parallel corpus to improve the word-word alignment learning. Consider the above discussed wrongly translated Hindi sentence,
**Hindi** : गर्मी से लू लगने से सिर दर्द तथा भूख न लगना.
*{garmi se loo lagne se sir dard tadha bhookh na lagna}*
*{Headache and lack of appetite because of sunstroke in summer.}*
After the cleaning process, the Hindi sentence has been correctly translated into English as,
**Correct English Translation :** *Headache and lack of appetite because of sunstroke in summer.*
MT system was able to generate good quality translations after training with the cleaned corpus. The quality of the translation has improved to more than 40% compared to SMT system with uncleaned corpus as shown in Table 2, 3, 4 and 5. We observed that the system fails to handle the rich morphology and we started investigating various ways to handle the morphological inflections.

### 3.3 SMT system with Suffix splitted corpus

We conducted experimenting with suffix splitting of the inflected words to handle the morphology. Consider a Hindi sentence,
फोटो व वीडियो कलाओं का प्रदर्शन और साथ में कई अन्य गतिविधियाँ।
*{photo va video kalavom ke pradarshan our saath men kayi anya gathividhiyaam}*
*{ Photo, video art and along with it many other activities}*
**Hindi sentence with suffix split:**
फोटो व वीडियो कला ओं का प्रदर्शन और साथ में कई अन्य गतिविधि याँ।
*{photo va video kala om ke pradarshan our saath men kayi anya gathividhi yaam}*

We have done the experiments after splitting the suffixes for the inflected words in the entire corpus. We have analyzed that, even though the suffixes are getting splitted, it leads to an increment in the alignment options and hence the quality of the translation is not improving to a great level. The quality of the translation has improved slightly more than Baseline SMT system as shown in Table 2, 3, 4 and 5. We have decided to experiment with IndoWordnet synset words to handle the vocabulary differences and ambiguity, after a detailed error analysis.

### 3.4 SMT system with IndoWordnet extracted words

The bilingually mapped words with its semantic and lexical relations with size of 200000 were extracted from Indowordnet [16]. We have generated parallel entries of words by considering all the possible synset word mappings for a single word. Consider the word *abandon* and it's generated synset wordmappings from IndoWordnet.

 *abandon:* स्थान_त्यागना स्थान_खाली_करना स्थान_छोड़ना
*{abandon:sthan_tyagna sthan_khali_karna sthan_chodna}{ abandon: abandon abandon abandon }*

The extracted Indowordnet synset words were augmented into the training corpus. Then the results were compared against the baseline system as shown in the Table 2, 3, 4 and 5. We have observed that the augmentation of synset words into training corpus not only helped in improving the quality of translation but also it helped in handing the lexical and semantic ambiguity as well. We further analyzed that the resultant translation fails to handle the various inflected forms, case markers etcetera at various times. Hence, we have decided to construct parallel entries of function words.

### 3.5 Corpus with Function words

We have prepared 15000 parallel entries of function words; suffix pairs etc over a period of 5 months and augmented it into the training corpus. Consider a sample English-Hindi Function word pair,

*Somebody : किन्ही-किन्ही लोगों {Somebody : kinhi-kinhi logon}*
*{ Somebody : Somebody }*

We observed that the grammatical structure as well as the quality of the translation has improved a lot after augmenting the corpus with Function words. The comparative results are shown in the Table 2, 3, 4 and 5. We further observed that even though the quality and structure of the translation is improving, the system fails to handle the verbal inflections properly. Hence, we started a study on Hindi verbal inflections.

### 3.6 Corpus with verb phrases

We have prepared and validated 85000 entries of English-Hindi verb phrases over a period of 6 months, which contain plentiful examples of various verbal inflections. Then we have augmented these verb phrases into the training corpus. Consider a sample verb phrase entry from the training corpus,

*Blow out of the water: भौंचक्का_होना {bhaimchakka hona}*

After analyzing the results, we have observed that the MT system was able to translate the verb phrases correctly to a great extent. The error analysis study shows that the quality of the translation has improved a lot and the results are shown in Tables 2, 3, 4 and 5.

## 4. Evaluation & Error Analysis

| English-Hindi Statistical MT Baseline System | | BLEU score | MET-EOR | TER |
|---|---|---|---|---|
| With Uncleaned Corpus | Without Tuning | 8.06 | 0.137 | 93.48 |
| | With Tuning | 12.76 | 0.144 | 91.94 |
| With CleanedCorpus | Without Tuning | 27.97 | 0.280 | 63.05 |
| | With Tuning | 29.56 | 0.293 | 60.92 |
| With Suffix Split Corpus | Without Tuning | 30.01 | 0.301 | 58.05 |
| | With Tuning | 31.21 | 0.311 | 55.12 |
| Corpus with Wordnet | Without Tuning | 37.31 | 0.363 | 43.91 |
| | With Tuning | 39.68 | 0.378 | 41.02 |
| Corpus with Function Words | Without Tuning | 43.55 | 0.412 | 37.89 |
| | With Tuning | 45.67 | 0.407 | 35.09 |
| Corpus With Verb Phrases | Without Tuning | 52.59 | 0.481 | 30.06 |
| | With Tuning | 55.87 | 0.492 | 28.23 |

Table 2: Results of English-Hindi SMT BLEU score, METEOR, TER Evaluations

We have used a tuning (MERT) corpus of 500 sentences as shown in Table 1. We have tested the translation system with 1000 sentences taken from the 'ILCI tourism, health' corpus as shown in Table 1. We have evaluated the translated outputs of both Hindi to English and English to Hindi SMT systems in all 5 categories. We have used various evaluation methods such as subjective evaluation, BLEU score (Papineni et al., 2002), METEOR and TER (Agarwal and Lavie 2008) to analyze better.

| Hindi-English Statistical MT Baseline System | | BLEU score | METEOR | TER |
|---|---|---|---|---|
| With Uncleaned Corpus | Without Tuning | 9.28 | 0.131 | 92.32 |
| | With Tuning | 11.91 | 0.139 | 91.11 |
| With Cleaned Corpus | Without Tuning | 28.03 | 0.210 | 63.54 |
| | With Tuning | 30.12 | 0.219 | 60.57 |
| With Suffix Split Corpus | Without Tuning | 32.07 | 0.235 | 66.32 |
| | With Tuning | 34.67 | 0.243 | 63.97 |
| Corpus with Wordnet | Without Tuning | 41.51 | 0.358 | 53.75 |
| | With Tuning | 43.56 | 0.381 | 51.21 |
| Corpus with Function Words | Without Tuning | 48.21 | 0.416 | 45.19 |
| | With Tuning | 51.43 | 0.423 | 42.17 |
| Corpus with Verb Phrases | Without Tuning | 58.67 | 0.563 | 35.32 |
| | With Tuning | 60.85 | 0.578 | 31.23 |

Table 3: Results of Hindi-English SMT BLEU score, METEOR, NER Evaluations

| English-Hindi SMT Baseline System | | Adequacy | Fluency |
|---|---|---|---|
| With uncleaned corpus | Without Tuning | 20.3% | 25.36% |
| | With Tuning | 22.8% | 31.1% |
| With Cleaned Corpus | Without Tuning | 59.34% | 67.34% |
| | With Tuning | 63.7% | 75.27% |
| With Suffix Split | Without Tuning | 64.10% | 76.78% |
| | With Tuning | 65.72% | 78.35% |
| Corpus with Wordnet | Without Tuning | 75.9% | 83.8% |
| | With Tuning | 77.13% | 85.65% |
| Corpus with Function Words | Without Tuning | 79.23% | 87.11% |
| | With Tuning | 81.21% | 89.22% |
| Corpus With Verb Phrases | Without Tuning | 86.43% | 92.70% |
| | With Tuning | 88.54% | 94.67% |

Table 4 : English-Hindi SMT Subjective Evaluation Results

| Hindi-English SMT Baseline System | | Adequacy | Fluency |
|---|---|---|---|
| With Uncleaned Corpus | Without Tuning | 17.56% | 24.67% |
| | With Tuning | 21.98% | 27.39% |
| With Cleaned Corpus | Without Tuning | 56.32% | 67.54% |
| | With Tuning | 59.78% | 74.34% |
| With Suffix Split | Without Tuning | 60.62% | 76.47% |
| | With Tuning | 62.98% | 77.76% |
| Corpus with Wordnet | Without Tuning | 72.39% | 85.18% |
| | With Tuning | 74.73% | 87.15% |
| Corpus with Function Words | Without Tuning | 78.93% | 89.64% |
| | With Tuning | 81.36% | 91.25% |
| Corpus With Verb Phrases | Without Tuning | 85.68% | 87.38% |
| | With Tuning | 88.01% | 90.39% |

Table 5: Hindi-English SMT System Subjective Evaluation Results

We have followed the subjective evaluation procedure as described in Sreelekha et.al.(2013) and the results are given in Table 4 and Table 5. The results of BLEU score, METEOR and TER evaluations are displayed in Tables 2 and 3. We have observed that, as the corpus is getting cleaned and more lexical resources are being used, the quality of the translation is improving. Hence, there is an incremental growth in adequacy, fluency, BLEU score, METEOR score and reduction in TER score. The fluency of the translation is increased up to 90.39% in the case of Hindi to English and up to 94.67 % in the case of English to Hindi, which is 4 times more than the baseline system results.

## 5. Conclusion

In this work, we have investigated on various ways to improve the quality of machine translation in a resource poor language Hindi. To improve the quality of translation, we have prepared and experimented with various lexical resources such as lexical words, function words, and verb phrases etcetera. We have discussed the six categories of experiments on top of the baseline phrase based SMT system with 24 trained models and its comparative performance in detail for both English–Hindi and Hindi-English pairs. The resultant SMT systems were able to handle the morphological infections and grammatical structures to a great extend. We have used various measures such as BLEU Score, METEOR, TER, subjective evaluation in terms of Fluency and Adequacy. Evaluation results show that there is an incremental growth for both English-Hindi and Hindi-English systems in terms of BLEU-Score, METEOR, Adequacy and Fluency. There is a gradual reduction in TER evaluation scores, which shows the improvement in translation quality. Our future work will be focused on investigating more lexical resources for improving the quality of Statistical Machine Translation systems for various language pairs.

**Acknowledgments**

This work is funded by Department of Science and Technology, Govt. of India under Women Scientist Scheme- WOS-A with the project code- SR/WOS-A/ET-1075/2014.